\documentclass{article}
\usepackage{spconf}
\usepackage{amsmath,epsfig}
\usepackage{amsfonts}

\usepackage{graphicx,color}
\graphicspath{{images/}}
\usepackage{verbatim}
\usepackage{url}
\usepackage[noadjust]{cite}
\usepackage{relsize} 
\usepackage{float}
\usepackage{caption}
\usepackage{subcaption}
\usepackage{verbatim}
\usepackage{bm}
\usepackage{rotating}
\usepackage{tikz}
\usepackage{mathtools} 
\usepackage{algorithm}
\usepackage{algorithmic}
\usepackage{dsfont}

\def\zZ{{\mathbb Z}}

\def\ind{{\mathds{1}}}

\newcommand{\myc}{\mathcal{C}}

\newcommand*{\myWidth}{135px}
\newcommand*{\myHeight}{121px}

\title{Estimation of Bandlimited Grayscale Images from the Single Bit Observations of Pixels Affected by Additive Gaussian Noise}
\name{Abhinav Kumar and Animesh Kumar}
\address{Department of Electrical Engineering\\
Indian Institute of Technology Bombay\\
Mumbai, India -- 400076\\
$[$abhinavkumar,animesh$]@$ee.iitb.ac.in}
\begin{document}
%
\maketitle
\begin{abstract}
The estimation of grayscale images using their single-bit zero mean Gaussian noise-affected pixels is presented in this paper. The images are assumed to be bandlimited
in the Fourier cosine transform (FCT) domain. The images are oversampled over their Nyquist rate
in the FCT domain. We propose a non-recursive approach based on first order approximation of Cumulative Distribution 
Function (CDF) to estimate the image from single bit pixels which itself is based on Banach's contraction theorem. 
The decay rate for mean squared error of estimating such images is found to be independent of the precision of
the quantizer and it varies as $O(1/N)$ where $N$ is the ``effective'' oversampling 
ratio with respect to the Nyquist rate in the FCT domain.
\end{abstract}
\section{Introduction}
\label{sec:intro}

We explore the image estimation from their noise-affected single bit pixels in this paper. There have been a few attempts to address this problem. 
Optics based approach \cite{hostettler15dispersion} has been used to recreate the images from one or two binary images.
However, Banach's Contraction Theorem \cite{kreyszig1989introductory},
\cite{brooks2009contraction} or linearisation based approach is not used. Also, no previous work gives a bound on the distortion (mean-squared 
error between estimated and original image)  of the reconstructed images.

There have been a few works in the area of 1-D signal processing. Reconstruction of  
1-D continuous signals has been done from the signed noisy samples using a random process as dither \cite{masry1981reconstruction}. 
A mean squared error of $O(1/N^{2/3})$ was obtained where $N$ is the oversampling factor with respect to the Nyquist rate.  
One dimensional continuous bandlimited signals were recovered using Picard's Iterations 
when such signals were ``companded'' by an another signal \cite{landau1961recovery}. The ``quantizer 
precision indifference principle'' of the distortion encountered while reconstructing Zakai-class \cite{zakai1965band} bandlimited 1-D signals 
from noisy samples has been carried out \cite{kumar2012estimation},\cite{kumar2013estimation}. The 
distortion 
was found to 
vary as $O(1/N)$ and was independent of the precision of the quantizer
i.e. the presence of noise and precision of the quantizer only decides the proportionality constant of the distortion. This work 
essentially extends their recursive scheme to images. Their scheme is also simplified based on the CDF linearisation. Also the filtering is carried out in FCT 
domain instead of the traditional Fourier domain.

The key contribution of the paper is as follows. If $N$ is the ``effective'' oversampling factor with respect to the 
Nyquist rate of a grayscale image in the FCT domain, then
a distortion of $O(1/N)$ can be achieved irrespective of the quantizer precision.

\section{Background}
\subsection{Fourier Cosine Transform}
The 1-Dimensional FCT and 1-D Inverse Fourier Cosine Transform (IFCT) \cite{rao2014discrete} for even signal $f(t)$ are defined by (\ref{eq:1dFCT})
\begin{equation}
F(\omega)=2\int\limits_{0}^{\infty}f(t)\cos(\omega t)dt; f(t)=\dfrac{2}{\pi}\int\limits_{0}^{\infty}F(\omega)\cos(\omega t)d\omega
\label{eq:1dFCT}
\end{equation}

\begin{figure}[htb]
\begin{minipage}[b]{.32\linewidth}
  \centering
  \scalebox{0.45}{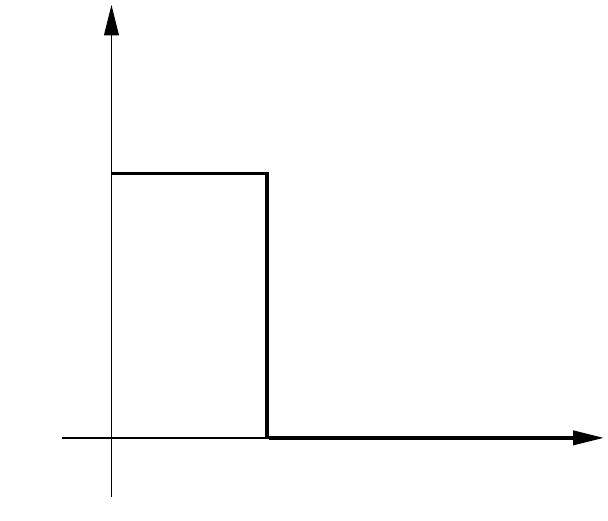}
  \centerline{(a)Ideal}\medskip
\end{minipage}
\hfill
\begin{minipage}[b]{0.32\linewidth}
  \centering
   \scalebox{0.45}{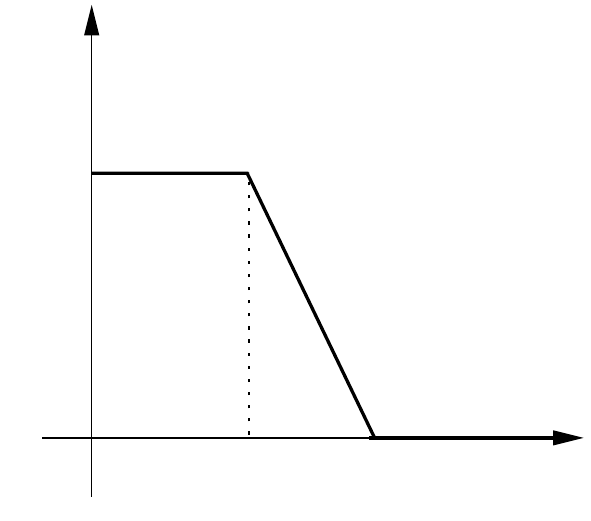}
  \centerline{(b)Tapered}\medskip 
\end{minipage}
\hfill
\begin{minipage}[b]{0.32\linewidth}
  \centering
  \centerline{\includegraphics[width=75px,height=75px]{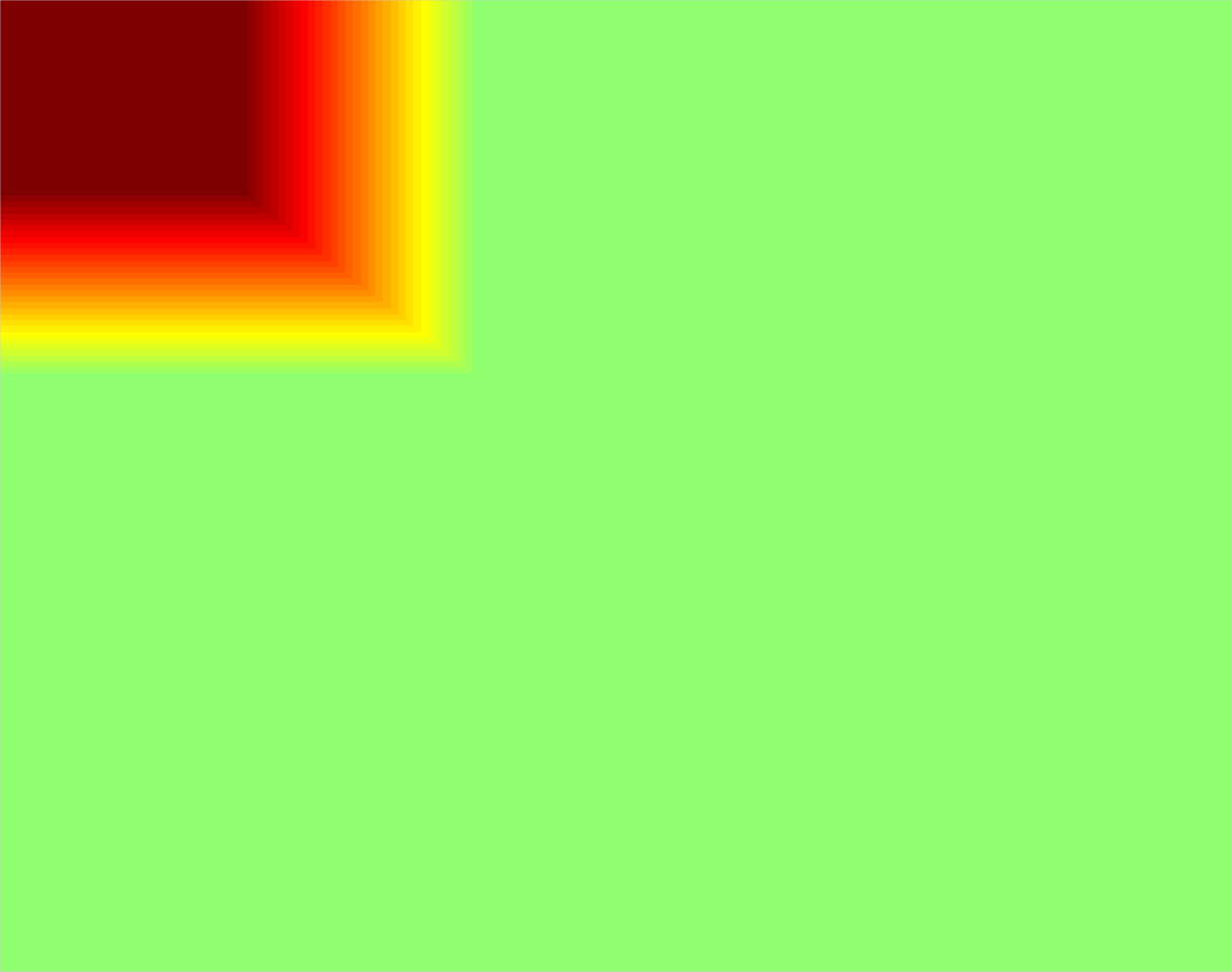}}
  \centerline{(c)Tapered Box,$\lambda=2$}\medskip 
  \label{fig:taperedBox}
\end{minipage}
\caption{Low Pass Filters in frequency domain.}
    \label{fig:zakai} 
\end{figure}

The interpolation of samples in time domain happens to be a low-pass filtering operation in frequency-domain. 
The interpolation kernel for the signal samples is then  $
f(t) =  \frac{2}{\pi} \int\limits_{0}^{\omega_m}1.\cos(\omega t)d\omega = \frac{2}{\pi} \frac{\sin(\omega_m t)}{t} 
$, which is a sinc function. 
Thus, the interpolation function for the FCT is same as the interpolation function for the Fourier Transform.
This kernel is not absolutely summable and hence we go for a bandlimited absolutely summable kernel. The FCT of the kernel $\phi(t)$ is shown in 
Fig. \ref{fig:zakai}. Clearly, 
for continuous frequency response 
$\Phi(\omega)$, we should have $\lambda > 1$.
The closed form expression of $\phi(t)$ \cite{kumar2012estimation} is then given by 
\begin{align}
 \phi(t) &=
  \begin{cases}
   1 + \frac{a}{\pi}        &,t = 0 \\
   \frac{\sin((\pi+a)t)\sin(at)}{at^2}        & ,\text{otherwise}\\
  \end{cases}
\end{align}
where $a = \frac{\lambda-1}{2}$.

Images have been shown to be better bandlimited in Cosine Transform Domain than in Fourier Transform Domain 
\cite{gonzalez2002digital},\cite{jain1989fundamentals}. Most of the energy 
of the images are contained in the upper left corner of the images \cite{ahmed1974discrete}.
The filter to be used should be a product of two $\phi$ kernels. The closed form expression of $\phi(x,y)$ is given by $\phi(x,y) = \phi(x)\phi(y) $. The filter $\phi(x,y)$
acts as a low pass filter for images. The frequency response of $\phi(x,y)$ denoted by $\Phi(\omega_1,\omega_2)$ is 
shown in the Fig. \ref{fig:zakai}c.

\subsection{Image and Noise Model}
We consider images $g(x,y)$ to be bandlimited in Zakai sense \cite{zakai1965band} in FCT domain i.e., there exists a $\phi(x,y;\omega_m,\omega_m)$ 
such that $ g(x,y) * \phi(x,y;\omega_m,\omega_m) = g(x,y)$
where $*$ denotes the linear convolution operation. Here, $\phi(x,y)$ is called the 
kernel function and  $(\omega_m,\omega_m)$ is the cutoff frequency in 
FCT domain. The other requirement is that image is bounded. In other words, $|g(x,y)| \le 1$. The image is corrupted by a zero mean additive Gaussian noise $W$ of 
variance  $\sigma^2$.
 
\subsection{Sampling Model}
\vspace{-0.7cm}
\begin{figure}[!htb]
\begin{center}
\scalebox{0.56}{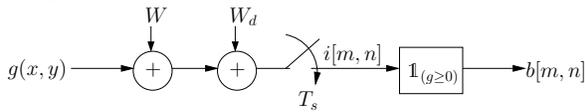}
\end{center}
\caption{\label{fig:samp2}Sampling Scheme for single bit precision pixels.}
\end{figure}
The indicator function models the single bit thresholding operation of the image pixels in Fig. \ref{fig:samp2}. $W_d$ and $T_s$ denote the 
dither noise and the sampling rate respectively. We also 
generalise the concept of Nyquist sampling rate from Fourier transform to FCT domain.
Let $f_m$ be the maximum frequency content of a 1-D signal in FCT domain.
Then, the sampling rate for perfect reconstruction $> {2f_m}$. If we oversample a signal $N$ times its Nyquist rate along each of the axes, 
then the sampling rate is ${N 2f_m}$.

Distortion measure $D$ considered for the images is mean-squared error i.e. 
$D = \frac{1}{M}\sum\limits_x \sum\limits_y [\widehat{g}(x,y) - g(x,y)]^2 $
, where $g(x,y)$ and $\widehat{g}(x,y)$ denote the original and estimated images respectively
and $M$ is total number of pixels in the image.

\section{Image Estimation Algorithms}
This section contains the image estimation algorithms using full precision as well as binary pixels.
\subsection{Estimating Bounded Zakai Sense bandlimited Images from Full Precision Pixels}
We first interpolate the noisy sampled pixels $i[m,n]$ (Fig. \ref{fig:samp2}) to get the image 
	\begin{equation}
	h(x,y)=\mathlarger{\mathlarger{‎‎\sum}}_{m\in\zZ} \mathlarger{\mathlarger{‎‎\sum}}_{n\in\zZ} i[m,n] \phi\left(\frac{x-mT_s}{T_s},\frac{y-nT_s}{T_s}\right)
	\end{equation}
Then, the estimate of the original image $g(x,y)$ is obtained by low passing the image $h(x,y)$ as 
	\begin{equation}
	   \widehat{g}(x,y) =  h(x,y) * \phi(x,y;\omega_m,\omega_m)
	\end{equation}
\subsection{Estimating Bounded Zakai Sense bandlimited Images from Binary Pixels}
If a constant signal $c$ corrupted by Gaussian noise with mean zero and variance $\sigma^2$ is observed through single bit quantisers, the mean of the observations converges to
$\myc(c)$ \cite{kumar2012estimation} where  $\myc(t)$ denote the CDF of the Gaussian noise. 

Here, we are given the noisy binary pixels $b[m,n]$ (Fig. \ref{fig:samp2})  
sampled at above the Nyquist rate  with values as 1 or 0 and thresholded at 0. The sampling rate $(T_s)$, 
oversampling factor $(N)$, steepness factor $(\lambda)$ and variance of the zero mean Gaussian noise $(\sigma^2)$
as well as the dither $(\sigma_d^2)$ are assumed to be known. 
Let $\myc(t)$ denote the CDF of the Gaussian random variable $\mathcal{N}(0,\sigma^2 + \sigma_d^2)$.
\begin{figure}[!htb]
  \centering
          \includegraphics[width=0.34\textwidth]{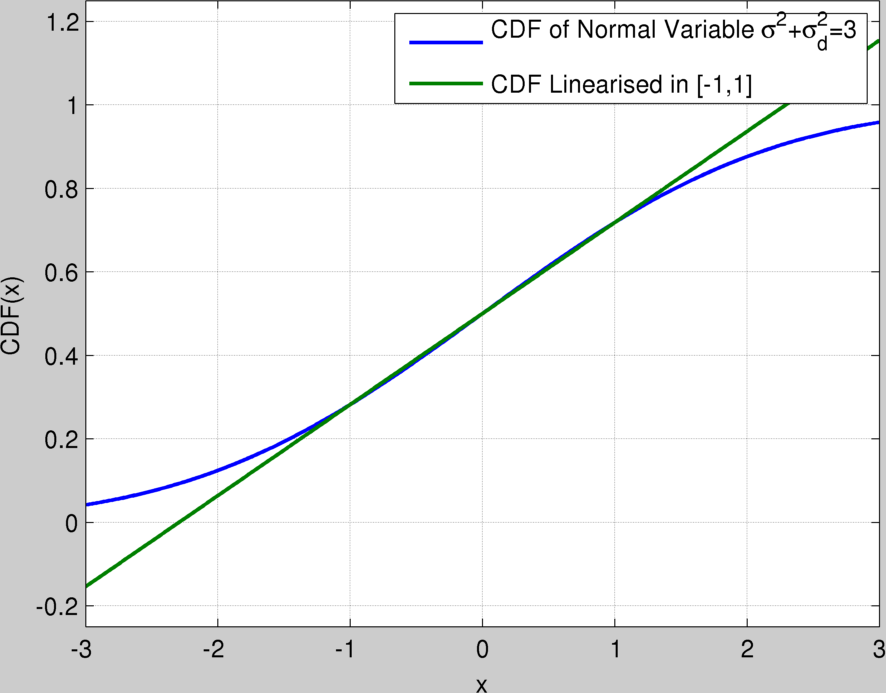}
    \caption{CDF of zero mean Gaussian random variable and its linearised approximation around [-1,1].}
    \label{fig:cdfGaussian}
\end{figure}

Interpolation of binary pixels gives us $H_N(x,y)$. Following thw procedure of statistical indifference\cite{kumar2012estimation}, it can be shown to 
converge to $\myc(g(x,y))*\phi(x,y;\omega_m,\omega_m)$ which is a non-linear estimate of $g(x,y)$. 
If a dither Gaussian noise of large variance is added before sampling, the CDF of the Gaussian random variable is nearly linear 
in the range $[-1,1]$. Fig. \ref{fig:cdfGaussian} shows the plot of CDF of 
Gaussian random variable and its first order approximation in the region $[-1,1]$. Clearly, as the variance of the Gaussian random variable increases,
the approximation of CDF by a straight line becomes more and more perfect.  Hence, the estimate is of the form $\alpha g(x,y)+\beta$.
\begin{figure*}[!htb]
\begin{minipage}[b]{.32\linewidth}
  \centering
  \centerline{\includegraphics[width=\myWidth,height=\myHeight]{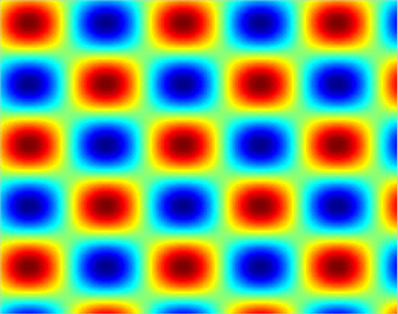}}
  \centerline{(a) Cosine image}\medskip
\end{minipage}
\hfill
\begin{minipage}[b]{.32\linewidth}
  \centering
  \centerline{\includegraphics[width=\myWidth,height=\myHeight]{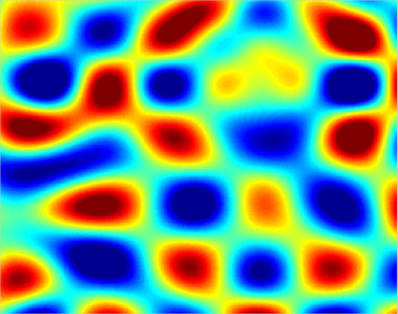}}
  \centerline{(b) Estimated Unquantised Pixels ${N = 1}$}\medskip
\end{minipage}
\hfill
\begin{minipage}[b]{.32\linewidth}
  \centering
  \centerline{\includegraphics[width=\myWidth,height=\myHeight]{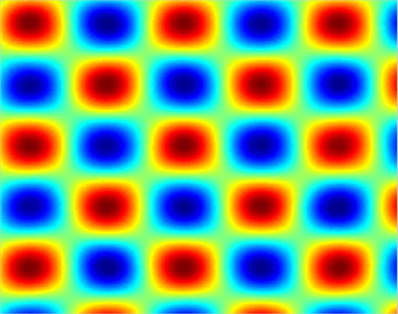}}
  \centerline{(c) Estimated Unquantised Pixels ${N = 32}$}\medskip
\end{minipage}
\begin{minipage}[b]{.32\linewidth}
  \centering
  \centerline{\includegraphics[width=\myWidth,height=\myHeight]{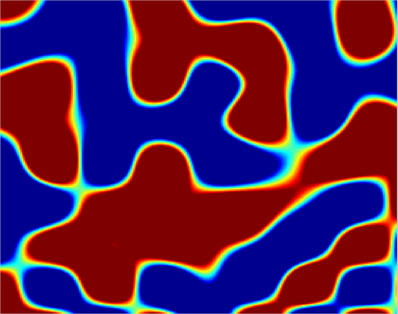}}
  \centerline{(d) Estimated Single Bit Pixels ${N = 1}$}\medskip
\end{minipage}
\hfill
\begin{minipage}[b]{.32\linewidth}
  \centering
  \centerline{\includegraphics[width=\myWidth,height=\myHeight]{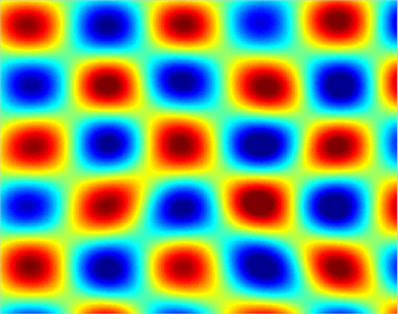}}
  \centerline{(e) Estimated Single Bit Pixels ${N = 32}$}\medskip
\end{minipage}
\hfill
\begin{minipage}[b]{.32\linewidth}
  \centering
  \centerline{\includegraphics[width=\myWidth,height=\myHeight]{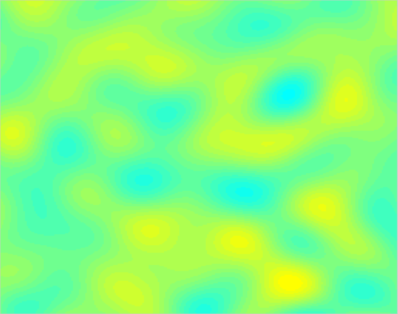}}
  \centerline{(f) Difference of (a) and (e)}\medskip
\end{minipage} 
    \caption{Estimation of synthetic Cosine image using Unquantised and Single Bit Pixels.}
    \label{fig:2048Full}
\end{figure*}
\begin{figure*}[!htb]
\begin{minipage}[b]{.32\linewidth}
  \centering
  \centerline{\includegraphics[width=\myWidth,height=\myHeight]{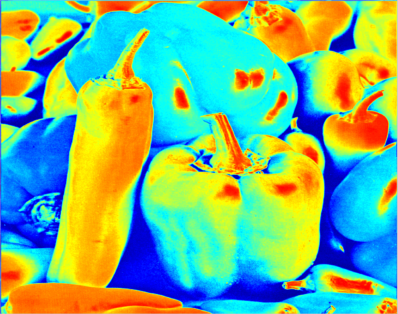}}
  \centerline{(a) Peppers image}\medskip
\end{minipage}
\hfill
\begin{minipage}[b]{.32\linewidth}
  \centering
  \centerline{\includegraphics[width=\myWidth,height=\myHeight]{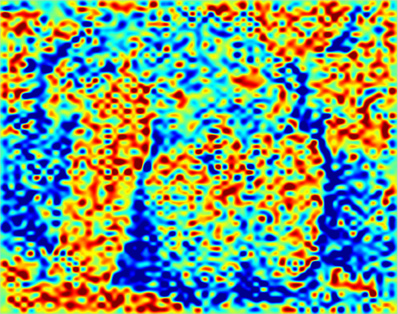}}
  \centerline{(b) Estimated Unquantised Pixels ${N = 1}$}\medskip
\end{minipage}
\hfill
\begin{minipage}[b]{.32\linewidth}
  \centering
  \centerline{\includegraphics[width=\myWidth,height=\myHeight]{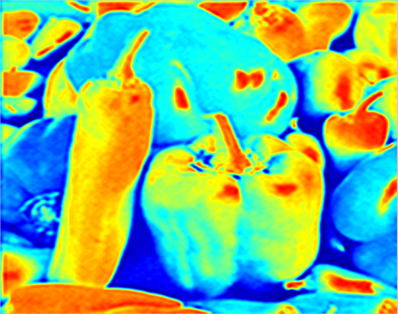}}
  \centerline{(c) Estimated Unquantised Pixels ${N = 32}$}\medskip
\end{minipage}
\begin{minipage}[b]{.32\linewidth}
  \centering
  \centerline{\includegraphics[width=\myWidth,height=\myHeight]{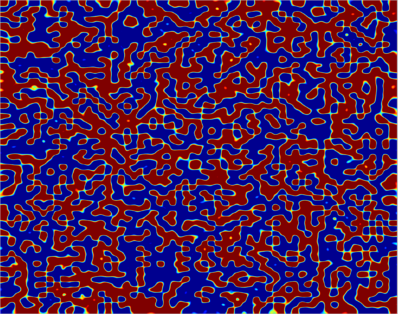}}
  \centerline{(d) Estimated Single Bit Pixels ${N = 1}$}\medskip
\end{minipage}
\hfill
\begin{minipage}[b]{.32\linewidth}
  \centering
  \centerline{\includegraphics[width=\myWidth,height=\myHeight]{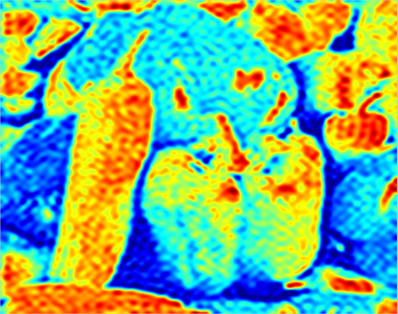}}
  \centerline{(e) Estimated Single Bit Pixels ${N = 32}$}\medskip
\end{minipage}
\hfill
\begin{minipage}[b]{.32\linewidth}
  \centering
  \centerline{\includegraphics[width=\myWidth,height=\myHeight]{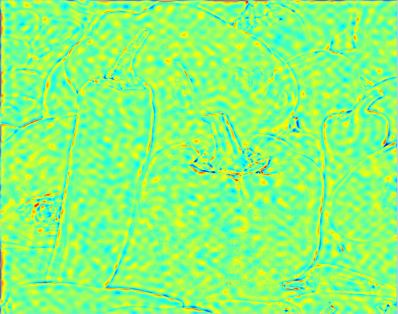}}
  \centerline{(f) Difference of (a) and (e)}\medskip
\end{minipage} 
    \caption{Estimation of Peppers image using Unquantised and Single Bit Pixels.}
    \label{fig:SingleDither=2.9}
\end{figure*}

Now, CDF can be directly inverted without any clipping or recursion to 
obtain the original image $g(x,y)$ back. Then, the direct 
algorithm \ref{al:5} is used to recover $g(x,y)$.

\begin{algorithm}[!htb]
\caption{Non-Recursive Image Denoising Algorithm using Single Bit Pixels}
\label{al:5}
\begin{algorithmic}
\REQUIRE $g(x,y)$
\STATE $H_N(x,y) \leftarrow \mathlarger{‎‎\sum}_{m\in\zZ}\mathlarger{‎‎\sum}_{n\in\zZ}\left(2b[m,n]-1\right)\phi\left(\frac{x-mT_s}{T_s},\frac{y-nT_s}{T_s}\right)*  \phi(x,y;\omega_m,\omega_m)$
\STATE $\alpha \leftarrow \frac{\myc(1)-\myc(-1)}{2}$
\STATE $\widehat{g}(x,y) \leftarrow \frac{H_N(x,y) }{2\alpha}$
\end{algorithmic}
\end{algorithm}

\section{Results}
We consider two images for our simulations- a synthetic Cosine image $Z = cos(2\pi f_mx)cos(2\pi f_my)$ and a real 
Peppers image. The synthetic image is already in the range $[-1,1]$ while the real image is 
initially converted to grayscale image in the range $[0,255]$ which is then
scaled in the range $[-1,1]$.
The number of DCT coefficients was $72\times 72$ for all simulations which corresponds to
$f_m=4$ Hz on both the axes. The steepness factor $\lambda$ has been kept at $2$ and the noise variance $\sigma^2$ for all images 
was $0.1$. The dither variance $\sigma_d^2$ was $2.9$. The spatial grid for $2048\times2048$ image for each of the axes has been kept as $-2.5575:.0025:2.56$ respectively. 
Denoising from single bit precision pixels was carried out using the non-recursive
algorithm \ref{al:5}.

Fig. \ref{fig:2048Full} and \ref{fig:SingleDither=2.9} shows the estimation of $2048\times 2048$ images with infinite and single bit 
quantizer precision for the two images respectively at different oversampling factors 
$N$ along each of the axis.

\begin{figure}[!htb]
    \centering    
    \includegraphics[width=240px,height=160px]{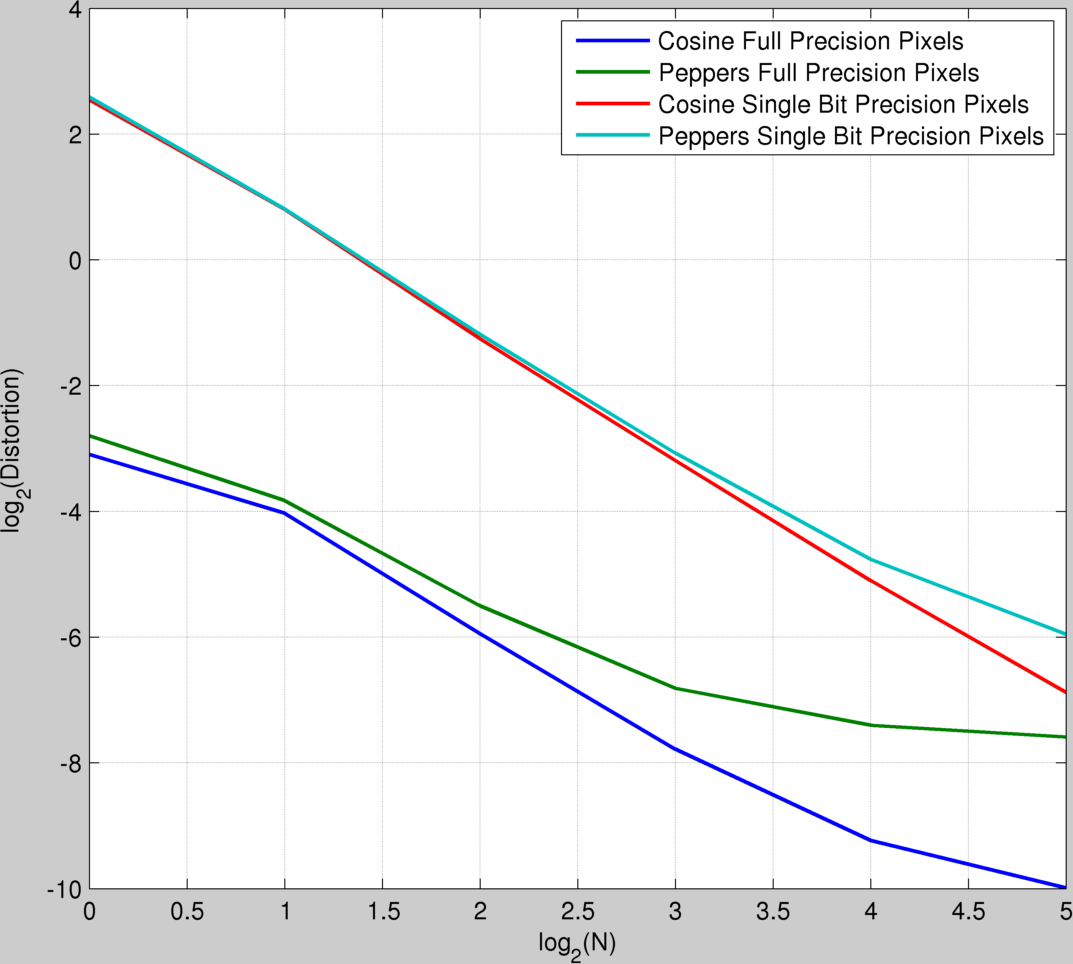}
    \caption{Plot of log of Distortion with log of oversampling factor along each axes $N$ for $2048 \times 2048$ images.}
    \label{fig:MSE2D2048}
\end{figure}

The plot of log of distortion for the two $2048\times 2048$ images versus log of the oversampling factor for the full precision and 
single bit precision case 
is shown in Fig. \ref{fig:MSE2D2048}.
\subsection{Discussions}
 The loss due to quantization is constant across oversampling factors for different images which is evident in Fig. \ref{fig:MSE2D2048}.
 The slope of the plot of log of distortion with log of oversampling factor for single bit precision samples in Fig. \ref{fig:MSE2D2048} is $-2$ which shows
 that distortion decreases as $O(1/N^2)$. The effective number of samples in the region is $N^2$ since we are oversampling each axis by $N$. Also, the distortion gap for unquantised 
 and quantised estimation is constant at different $N$. 
 
 The slope of the plot of log of distortion with log of oversampling factor for full precision samples 
 will become more accurate at the cost of computational resources.
 
 Some of the edges (which are high pass in nature) are not recovered back because of the low pass nature 
 of the kernel function. The distortion due to such out of band component $D_{out of band}$ 
 adds up linearly to the distortion caused due to
 reconstruction. Hence, for real images, we have $D_{observed}-D_{out of band}$ varies as $O(1/N)$.
\section{Conclusions}
The estimation of images from infinite bit precision and single bit pixels has been worked out for 2-D grayscale images. 
However, the results seem astonishing in the sense that an image can be estimated just by using the single-bit pixels. The 
tradeoff lies between the number of bits of quantizer precision and the oversampling introduced.
The distortion for the reconstruction varies as $O(1/N)$ for images 
which is independent of the precision of the quantizer where $N$ is the effective oversampling ratio.
 \section{Future Works}
Quantizer precision indifference principle can be extended to setups where basis is not ordered such as Wavelets and
 Contourlets. It would be interesting to see how this procedure behaves when we couple it with Compressing Sensing Algorithms 
 and the edge preserving priors.
 Estimation without using dither when the CDF is non-linear in the range $[-1,1]$ 
 would also be a good territory to explore.
 
\section*{Acknowledgments}
The authors would like to thank Nikunj Patel and Saurabh Kumar for discussions on this topic. 
Nikunj was also kind enough to
allow the simulations run on the TI-DSP lab machines. Saurabh was patient enough in listening to several problems.
\bibliographystyle{IEEEbib}
\bibliography{myBibliographyFile}

\end{document}